\documentclass[letterpaper, 10 pt, conference]{ieeeconf}  

\IEEEoverridecommandlockouts                              

\usepackage{graphics} 
\usepackage{epsfig} 
\usepackage{graphicx}
\usepackage{mathptmx} 
\usepackage{times} 
\usepackage{amsmath}
\usepackage{amssymb}
\usepackage{microtype}
\usepackage{algorithm, float}
\usepackage{algorithmic}
\usepackage{subfigure}
\usepackage{booktabs} 
\usepackage{hyperref}
\hypersetup{breaklinks=true,colorlinks=true,linkcolor=blue,citecolor=blue}
\usepackage{color}

\newcommand{\vect}[1]{\mathbf{#1}}

\DeclareMathOperator{\argmin}{arg\,min}
\DeclareMathOperator{\argmax}{arg\,max}

\DeclareMathOperator{\Prob}{\mathbb{P}}
\newtheorem{remark}{\textbf{Remark}}

\title{\LARGE \bf
Estimation and Planning of Exploration Over Grid Map Using A Spatiotemporal Model with Incomplete State Observations}

\author{Hyung-Jin Yoon$^{1}$, Hunmin Kim$^{2}$, Kripash Shrestha$^{1}$,  Naira Hovakimyan$^{2}$, and Petros Voulgaris$^{1}$
\thanks{*Research supported by NSF CPS \#1932529, NSF CMMI \#1663460, and UNR internal funding.}
\thanks{$^{1}$Hyung-Jin Yoon and Petros Voulgaris are with the Department of Mechanical Engineering, University of Nevada, Reno, NV 89557, USA
        {\tt\small \{hyungjiny, kshrestha, pvoulgaris\}@unr.edu}}%
\thanks{$^{2}$Hunmin Kim and Naira Hovakimyan are with the the Department of
Mechanical Science and Engineering, University of Illinois at Urbana-Champaign, Urbana, IL 61801, USA
        {\tt\small \{hunmin, nhovakim\}@illinois.edu}}%
}

\begin{document}

\maketitle
\thispagestyle{empty}
\pagestyle{empty}

\begin{abstract}
Path planning over spatiotemporal models can be applied to a variety of applications such as UAVs searching for spreading wildfire in mountains or network of balloons in time-varying atmosphere deployed for inexpensive internet service. A notable aspect in such applications is the dynamically changing environment. However, path planning algorithms often assume static environments and only consider the vehicle's dynamics exploring the environment. We present a spatiotemporal model that uses a cross-correlation operator to consider spatiotemporal dependence. Also, we present an adaptive state estimator for path planning. Since the state estimation depends on the vehicle's path, the path planning needs to consider the \emph{trade-off between exploration and exploitation}. We use a high-level decision-maker to choose an explorative path or an exploitative path. The overall proposed framework consists of an adaptive state estimator, a short-term path planner, and a high-level decision-maker. We tested the framework with a spatiotemporal model simulation where the state of each grid transits from normal, latent, and fire state. For the mission objective of visiting the grids with fire,  the proposed framework outperformed the random walk (baseline) and the single-minded exploitation (or exploration) path. 
\end{abstract}

\section{INTRODUCTION}

There are applications where planning with spatiotemporal models are useful. For example, path planning of UAV over mountains in Northern California can be modeled with a spatiotemporal model which can consider the dynamics of fire propagation and random initiation of fire~\cite{NBCNews}. Deployment and planning of the network of balloons for inexpensive wireless internet service~\cite{nagpal2017project} need to consider the dynamic atmosphere to keep the balloons in the designated altitude and horizontal positions. Also, path planning for mobile balloons equipped with a wind turbine to harvest wind-energy needs to consider how the wind is distributed over space and time~\cite{du2019energy}. Road condition grid map modeled as a spatiotemporal model was used to ensure the safety of autonomous cars by dynamically estimating the unknown road condition and using the state estimate for controlling the autonomous cars~\cite{kim2020robust}.

There are a number of papers on path-planning and coverage control over a map to cover or explore points of interest (contamination or wildfire). In~\cite{pham2018distributed}, the authors consider UAVs to cover dynamically changing fire-front of a wildfire simulation using a nonlinear control method to cover the area with a mobile sensing network~\cite{schwager2011eyes}. In~\cite{sung2019competitive}, a competitive tree search algorithm is employed in a UAV application where the UAVs explore the unknown source of contamination on the lake. In~\cite{leonard2007collective}, a multi-armed bandit problem is formulated for searching oil spills over the ocean. In the aforementioned applications, the authors assumed either complete state observation or the state of the environments are constant. The assumptions are not suitable for the tasks because the environment is dynamically changing and does not allow direct state observation. For example, the pollutant on the lake would flow to a different location over time. Also, UAVs do not see the entire map but only the area near the vehicles.

Exploration of unseen environments is an active topic in robotics. Simultaneous localization and mapping (SLAM) is a widely used problem formulation for exploration tasks. SLAM is essentially a state-estimation problem where the state includes the relative position between UAVs and landmarks or image features (edge, curvature, etc.). Despite the original formulation as state estimation, SLAM would not be a suitable approach for the path planning task with spatiotemporal model because SLAM techniques often focus on how efficiently to (fast) process camera images while assuming a static environment~\cite{mur2015orb}.
\begin{figure}[ht]
\begin{center}
\centerline{\includegraphics[width=\columnwidth]{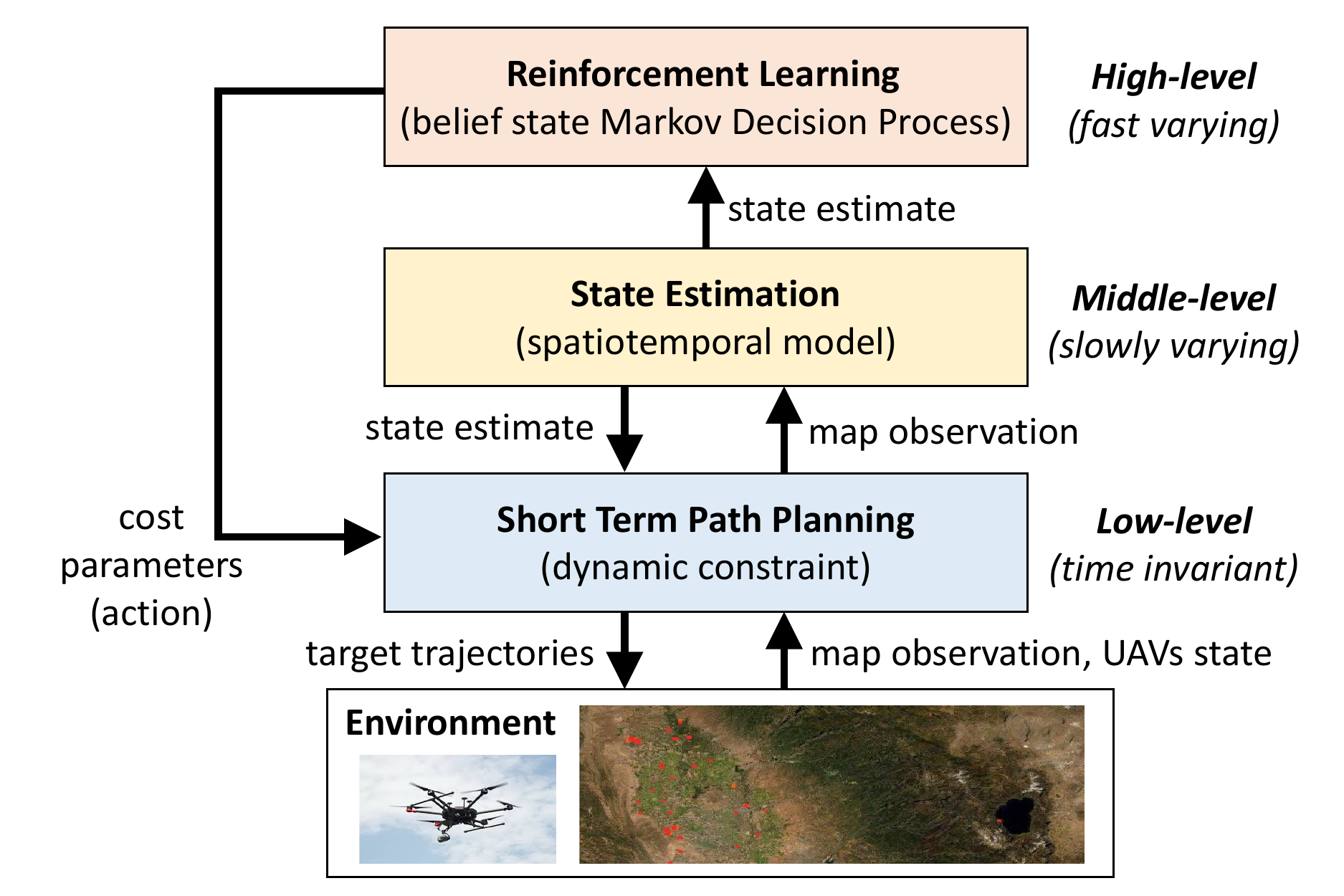}}
\caption{Multi-level adaptive path planning framework. The high-level decision-making module (reinforcement learning agent) chooses the path planning module's behavior to be explorative or exploitative.}
\label{fig/diagram}
\end{center}
\vskip -0.2in
\end{figure}

In this paper, we present an adaptive path planning framework that learns a spatiotemporal model to estimate hidden states of the dynamically changing grid map.  In contrast to the previous results that assume either  static environment or complete state observation for the path planning over dynamic environments, we consider state estimation, system identification, and exploration-exploitation trade-off for vehicle missions in unknown environments. As in Fig.~\ref{fig/diagram}, the proposed framework has a hierarchical structure similar to the structure having perception, planning, and control module in autonomous driving~\cite{serban2018standard}. We design the interface between modules in the multi-level structure. In the proposed framework, we embed a binary decision making on exploration or exploitation in the interface between high-level decision making (reinforcement learning agent) and the short-term planning module (sample-based path planner). To ensure convergence of the multiple components' adaptation, we employ multi-time scale stochastic optimization approach used in ~\cite{zhang2005tale, heusel2017gans}.

The remainder of this paper is organized as follows: In Section~\ref{sec:model}, we introduce a spatiotemporal model of two-dimensional grid map. In Section~\ref{sec:algorithms}, we describe the algorithms to estimate the state of the grid map and plan with the uncertain grid map environment. In Section~\ref{sec:numerical_examples}, a numerical example is presented. Finally, we give our concluding remarks in Section~\ref{sec:conclusion}.

\subsection*{Notation and terminology}
The following notation is adopted: $|{\cal S}|$ denotes the cardinality of the set for any finite set ${\cal S}$; ${\mathbb E}[\cdot]$ denotes the expectation operator; $\Prob[\cdot]$ denotes the probability of an event; $[\vect{x}]_i$ is its $i$-th element for any vector $\vect{x}$; $[\vect{P}]_{ij}$ indicates its element in $i$-th row and $j$-th column for any matrix $\vect{P}$; $[\vect{T}]_{ijk}$ indicates its element with the index $(i, j, k)$ for a three dimensional tensor $\vect{T}$;
$\mathbf{A}\odot\mathbf{B}$ denotes element-wise multiplication between tensors $\mathbf{A}$ and $\mathbf{B}$. We use upper case letter for random variable and lower case letter for its realization. The two dimensional (2D) cross-correlation operator\footnote{The 2D cross-correlation operator is the same operator as the convolution layers used in the convolutional neural networks (CNNs).} $\star$ is defined by $\vect{y} = \vect{k} \star \vect{x}$ $[\vect{y}]_{m,n} =  \sum_{j=-\infty}^{\infty}\sum_{i=-\infty}^{\infty}[\vect{x}]_{i,j}\cdot [\vect{k}]_{m-i, n-j}$, where the infinite series become finite series given the size of the kernel $\vect{k}$ and the size of the tensor input $\vect{x}$ such as grid maps or images.  For simple notations, we use the same tensor symbols after tensor dimension permutation, e.g., $\mathbf{u} \in \mathcal{P}^{|\mathcal{S}| \times H \times W}$ and $\mathbf{u} \in \mathcal{P}^{|H \times W \times \mathcal{S}| }$. We share the same symbol even if the estimates are calculated from different algorithms, e.g., we use $\hat{\mathbf{p}}$ for denoting the state estimate regardless of the methods (model-based or dynamic autoencoder). Other symbols and notations are listed in Table~\ref{tab:notation}.
\begin{table}[!t]
\caption{List of Notations}
\label{tab:notation}
\begin{center}
\begin{small}
\begin{tabular}{cl}
\toprule
Notation & Description \\
\midrule
$H, W$ & Height and width of the grid map \\ 

$\mathcal{S}$    & State set with number of elements $|\mathcal{S}|$\\

$\mathcal{O}$    & Observation set with number of elements $|\mathcal{O}|$\\

$k$    & Time index for adaptation and estimation \\ 

$\vect{X}_k$    & State of the grid map at $k$ \\

$\vect{Y}_k$    & Observation grid map at $k$ \\


$\vect{O}$ & Observation probability matrix in~\eqref{eq:observation_model}\\

$\Phi(\cdot)$ & State transition probability operator in~\eqref{eq:state_transition_operator} \\

\rule{0pt}{2ex}
$\mathcal{P}^{|\mathcal{S}| \times H \times W}$ & Set of probability distribution on the grid map\\
    
$\hat{\mathbf{p}}_k$ & State estimate of the grid map at time $k$\\

$\mathbf{u}_k$ & State predictor of the grid map at time $k$\\

$\hat{\mathbf{u}}_k$ & State predictor of dynamic autoencoder in~\eqref{eq:dynautoenc}\\


$t$    & Time index for path-planning\\ 

$\vect{z}_{t}$  & Position coordinate of the robot at time $t$\\

\bottomrule
\end{tabular}
\end{small}
\end{center}
\vskip -0.1in
\end{table}

\section{SPATIOTEMPORAL MODEL}\label{sec:model}
We consider the following spatiotemporal model defined with a finite state set $\mathcal{S}$ for each grid. Let $W$ and $H$ be the width and height of the grid map. For each grid, the set of possible state value is denoted as $\mathcal{S}$, i.e. $\mathcal{S}=\{ 1, 2, \dots, |\mathcal{S}|\}$. The state of the grid map $\vect{X}$ is defined as the following matrix
\begin{equation*}
    \vect{X}:=\left[\begin{array}{cccc}
x_{1,1}&x_{1,2}&\cdots &x_{1,W}\\
x_{2,1}&x_{2,2}&\cdots &x_{2,W}\\
\vdots & &\ddots &\vdots\\
x_{H,1}&x_{H,2}&\cdots &x_{H,W}\\
\end{array}\right],
\quad
x_{i,j} \in \mathcal{S}.
\end{equation*}

The state at each grid is observed as one of the finite observation sets, i.e. $\mathcal{O}=\{ 1, 2, \dots,|\mathcal{O}|\}$. The observation $y$ at a grid is associated with the state $x$ at the grid and is determined by the following probability:
\begin{equation}\label{eq:observation_model}
\mathbb{P}[y=l|x=m] = [\vect{O}]_{m,l}, \quad  \forall l \in \mathcal{O}, m \in \mathcal{S},
\end{equation}
where $\vect{O}$ denotes the observation probability matrix, which has strictly positive elements. Through the observation probability in~\eqref{eq:observation_model}, the output $\vect{Y}$ is determined given $\vect{X}$ with the following shape: 
\begin{equation*}
    \vect{Y}:=\left[\begin{array}{cccc}
y_{1,1}&y_{1,2}&\cdots &y_{1,W}\\
y_{2,1}&y_{2,2}&\cdots &y_{2,W}\\
\vdots & &\ddots &\vdots\\
y_{h,1}&y_{H,2}&\cdots &y_{H,W}\\
\end{array}\right],
\quad
y_{i,j} \in \mathcal{O}.
\end{equation*}
The spatiotemporal dependence along grids and time axis is considered in the following state transition probability model. Let $k$ denote the time index. The set of vertical and horizontal indices of the grid map is denoted as $I = \{1,2,\dots,H\}$ and $J = \{1,2,\dots,W\}$. The state transition model uses the 2D cross-correlation operator $\star$ to consider the spatial dependence as follows:
\begin{equation}\label{eq:cross-correlation}
    [\phi_{k+1}]_m = [\vect{b}]_m + \sum_{n=1}^{|\mathcal{S}|} [\vect{w}]_{n,m} \star [\phi_k]_m, \quad \forall m\in\mathcal{S},
\end{equation}
where $\phi_k\in\mathcal{P}^{|\mathcal{S}| \times H \times W}$ is initialized with initial belief state estimate, i.e. $\mathbb{P}\left\{[\vect{X}_0]_{i,j}=m \right\}, \, \forall m,i,j \in  \mathcal{S} \times I \times J$, and recursively updated with the current state predictor. In~\eqref{eq:cross-correlation},  $\vect{b}\in\mathbb{R}^{|\mathcal{S}|}$ denotes the bias and $\vect{w}\in\mathbb{R}^{|\mathcal{S}| \times |\mathcal{S}| \times H_{\vect{k}}\times W_{\vect{k}}}$ denotes the kernel of the 2D cross-correlation.
With the cross-correlation operation defined in~\eqref{eq:cross-correlation}, we calculate the state probability in the next time step as follows: 
\begin{equation}\label{eq:normalization}
    \mathbb{P}\left\{[\vect{X}_{k+1}]_{i,j}=m \right\} = [\phi_{k+1}]_{m,i,j} / \sum_{n=1}^{|\mathcal{S}|} [\phi_{k+1}]_{n,i,j},
\end{equation}
where $[\phi_{k+1}]_m$ is calculated using~\eqref{eq:cross-correlation}.

\begin{figure}[ht]
\begin{center}
\centerline{\includegraphics[width=\columnwidth]{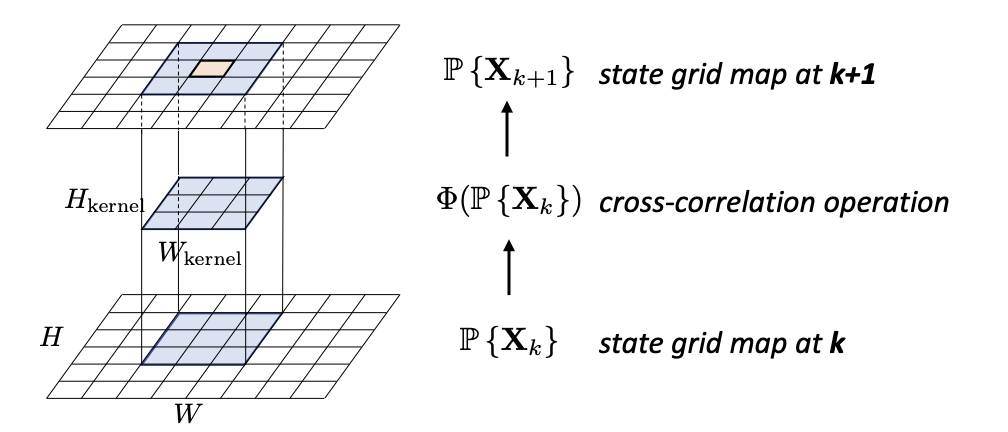}}
\caption{State transition model using the cross-correlation operator in~\eqref{eq:state_transition_operator}. The state probability of the grid colored light red at time $k+1$ depends on the state probability of the neighboring grids at time $k$ covered by the kernel colored light blue.}
\label{fig/state_transition_diagram}
\end{center}
\vskip -0.2in
\end{figure}

For an ease of notation, let us define the following operator $\Phi: \mathcal{P}^{|\mathcal{S}| \times H \times W} \rightarrow \mathcal{P}^{|\mathcal{S}| \times H \times W}$ as:
\begin{equation}\label{eq:state_transition_operator}
\begin{aligned}
    &\mathbb{P}\left\{[\vect{X}_{k+1}]_{i,j}= m, \,
    \forall m,i,j \in  \mathcal{S} \times I \times J \right\} \\
    &= \Phi\left(\{\mathbb{P}\left([\vect{X}_k]_{i,j}=m \right), \, \forall m,i,j \in  \mathcal{S} \times I \times J \}\right),
\end{aligned}
\end{equation}
where the calculation follows from~\eqref{eq:cross-correlation} and~\eqref{eq:normalization}. The cross-correlation operation is illustrated in Fig.~\ref{fig/state_transition_diagram}.

The spatiotemporal grid map model introduced in this paper uses a hidden Markov model (HMM) consisting of the state transition in~\eqref{eq:state_transition_operator} and the observation model in~\eqref{eq:observation_model}. We simulate the HMM by random sampling state values in grids using the state transition operator
defined in~\eqref{eq:state_transition_operator} and then again sampling the observation values in grid map using the observation probability given the sampled state values and the observation model in~\eqref{eq:observation_model}.

\begin{figure}[ht]
\begin{center}
\centerline{\includegraphics[width=0.8\columnwidth]{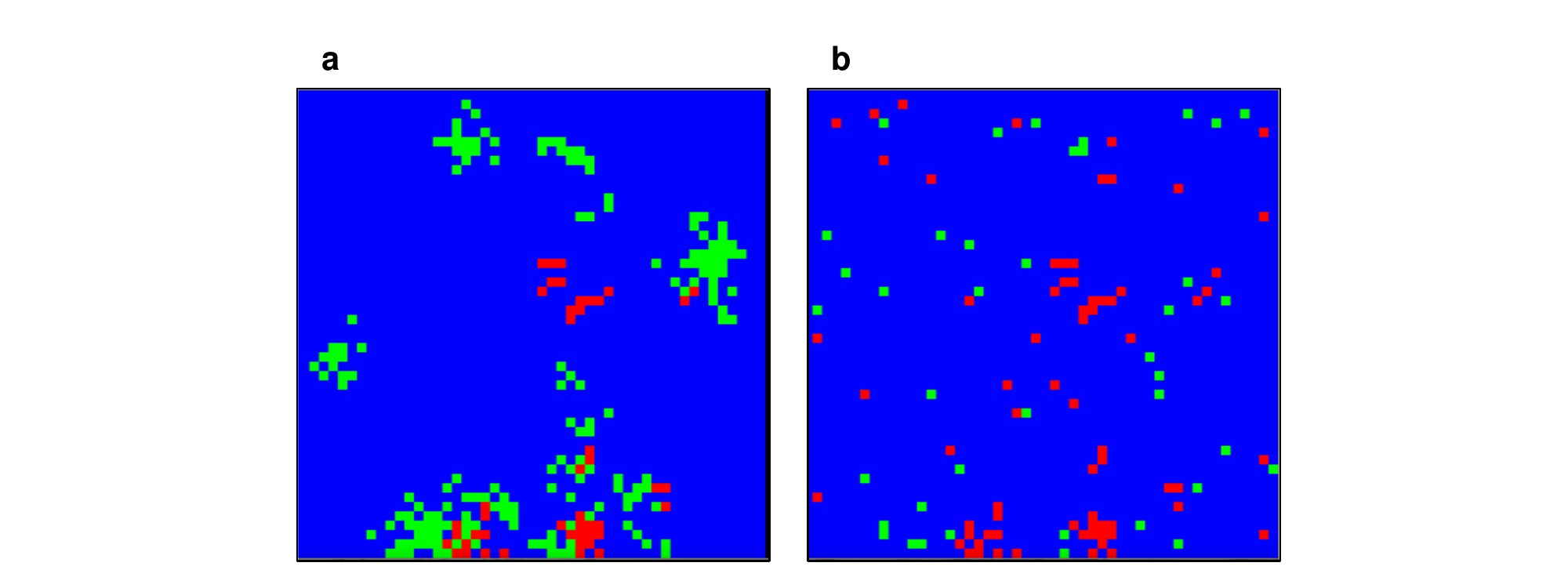}}
\caption{Incomplete state observation: (a) True state map. Each grid has three possible states(normal:blue, latent:green, abnormal:red). (b) Observed state map. See the illustrative video lined at 
\url{https://youtu.be/fTMNHY6JU8k}.}
\label{fig/state_observation}
\end{center}
\vskip -0.2in
\end{figure}

We present a simulation of the model to illustrate its use. In this simulation, we consider the three possible states (0: normal, 1: latent, 2: abnormal). Similar to typical Markov chains, the state transition follows in the order that starts with the normal state, becomes cautious (but also latent), changes to the abnormal state, and finally returns to the normal state while randomly allowing other transitions. The state transition probability depends on the state of the neighboring grids as in Fig.~\ref{fig/state_transition_diagram}. In this simulation, we assume that the latent state is less likely to be observed, i.e.  the grids having the latent state will be observed as normal with 80\% chance, cautious with 15\% chance, and abnormal with 5\% chance. As shown in Fig.~\ref{fig/state_observation}b, the latent state (colored green) is mostly hidden (\emph{incomplete state observation}) in the observation grids map. Also, the 2\textsuperscript{nd} state (green) and the 3\textsuperscript{rd} state (red) are spatially clustered due to the spatial dependence in the cross-correlation operator.   

\section{ALGORITHMS}\label{sec:algorithms}
Our proposed estimation and planning with the hidden Markov model  consist of state estimation and path planning. In this section, when we describe the state estimation algorithms, and we increase the complexity as follows:~\ref{sec:BayesinEst}. \emph{Model Based State Estimation}  uses the recursive Bayesian state estimation assuming that the structure and parameter of the model are given;~\ref{sec:SysID}. \emph{Online Parameter Estimation} uses a dynamic autoencoder taking similar approaches in~\cite{mirowski2010dynamic, chung2015recurrent, hernandez2018variational}, where dynamic deep neural networks are trained to maximize likelihoods;~\ref{sec:PathSysId}. \emph{Path Dependent Estimation}  considers path-dependent observation, i.e. we can only observe the grids where a robot visited. In~\ref{sec:planning}. \emph{Short-Term Path Planning},   
we use a random search method similar to stochastic model predictive control (or sample-based path planning)~\cite{williams2017model}, based on the state estimate grid map from ~\ref{sec:PathSysId}. Finally, in~\ref{sec:RL}. \emph{Reinforcement Learning to Guide Short-Term Planning}, we integrate the state estimation and the path planning by considering the trade-off between exploration and exploitation.
\subsection{Model Based State Estimation}\label{sec:BayesinEst}
\begin{figure}[ht]
\vskip 0.0in
\begin{center}
\centerline{\includegraphics[width=0.8\columnwidth]{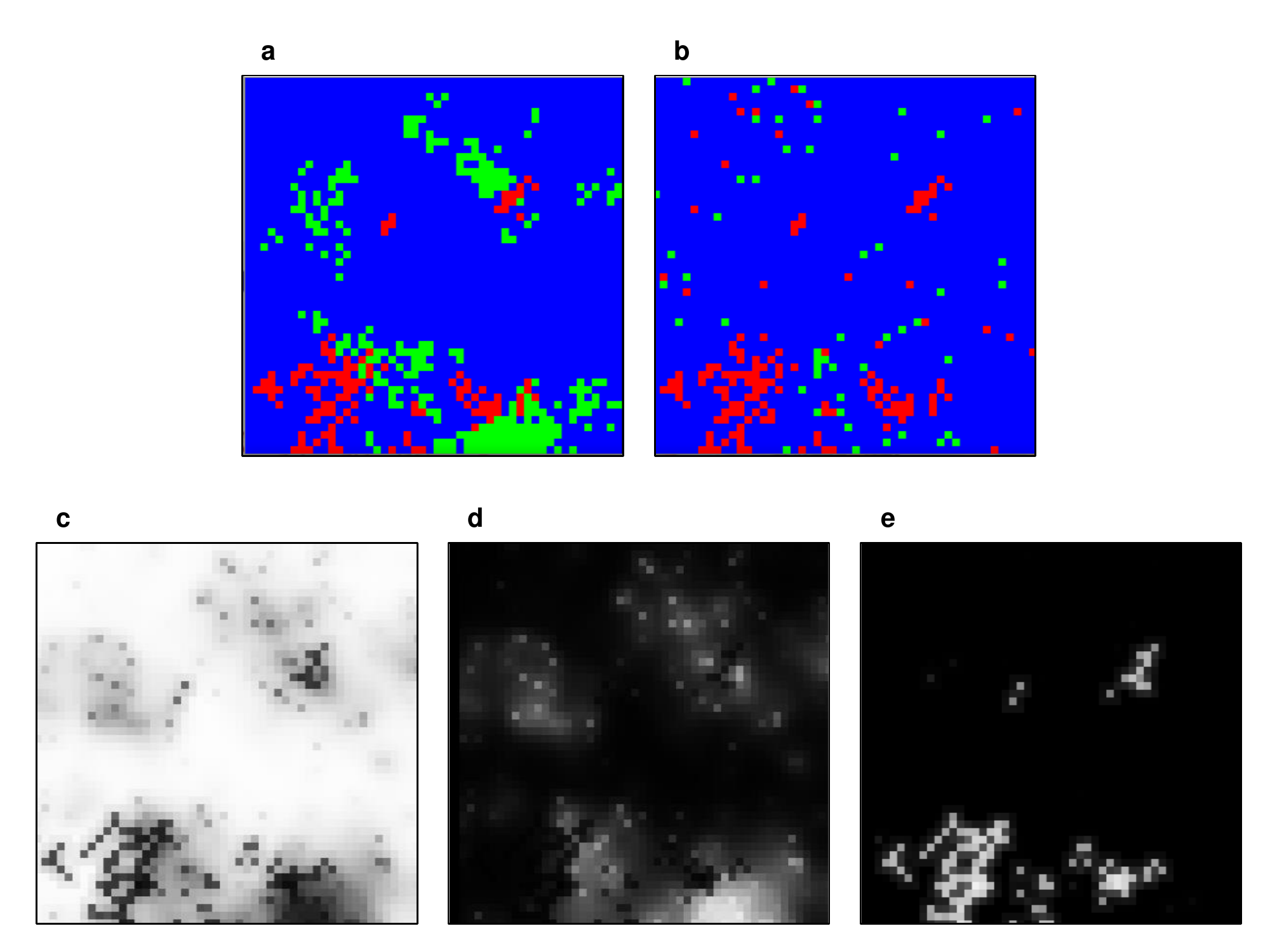}}
\caption{State estimation with known model: (a) True state map. Each grid has three possible states (blue/green/red) which are spatially and temporally dependent. (b) Observed state map. Observations of the green state are less likely than the others. Belief probability of (c) state 1 (blue); (d) state 2 (green); (e) state 3 (red). The light intensity of each pixel is proportional to the probability of the state estimate. See the illustrative video linked at
\url{https://youtu.be/ajCEbuEkpbY}.}
\label{fig/state_estimation}
\end{center}
\vskip -0.2in
\end{figure}

Here, we assume that the full model knowledge is available and only consider the uncertainties due to the noisy observation. The parameters of the spatiotemporal model include the observation probability matrix $\mathbf{O}\in\mathbb{R}^{|\mathcal{S}| \times |\mathcal{O}|}$ and the weight and bias of the cross-correlation operator. In the example shown in Fig.~\ref{fig/state_estimation}, the observation probability matrix $\mathbf{O}$ is only $3\times3$ matrix and the weight of the correlation operator $\mathbf{w}$ is $3\times3\times3\times3$ tensor and the bias $\mathbf{b}$ is a zero vector with three elements.

The hidden state can be estimated using a Bayesian recursive state estimation~\cite{chen2003bayesian}. The state estimate $\hat{\mathbf{p}}_k \in \mathcal{P}^{|\mathcal{S} | \times H \times W}$ is defined as follows:
\begin{equation*}
    \hat{\mathbf{p}}_k := \{[\hat{\mathbf{p}}_k]_{m,i,j} | \mathbb{P}\left([\vect{X}_{k}]_{i,j}= m | \mathcal{H}_k \right), \forall m,i,j \in  \mathcal{S} \times I \times J\},
\end{equation*}
where $\mathcal{H}_k$ denotes information collected until time $k$, i.e. $\mathcal{H}_k : = \{\vect{y}_1, \vect{y}_2, \dots, \vect{y}_k\}$.
The state predictor ${\mathbf{u}}_k \in \mathcal{P}^{|\mathcal{S} | \times H \times W}$ defined below
\begin{equation*}
    {\mathbf{u}}_{k+1} := \{[\hat{\mathbf{p}}_{k+1}]_{m,i,j} | \mathbb{P}\left([\vect{X}_{k}]_{i,j}= m | \mathcal{H}_k \right), \forall m,i,j \in  \mathcal{S} \times I \times J\}
\end{equation*}
is recursively updated from an initial value $\mathbf{u}_0 \in \mathcal{P}^{ H \times W \times |\mathcal{S}|}$ in the two steps:

(1) Calculate the state estimate using $\mathbf{y}_k$ and $\mathbf{u}_k$ as
\begin{equation}\label{eq:state_estimate_eqn}
    [\hat{\mathbf{p}}_k]_{i,j} = \frac{\mathbf{B}([\mathbf{y}_k]_{i,j}) [\mathbf{u}_k]_{i,j}}{\mathbf{b}([\mathbf{y}_k]_{i,j})^\top [\mathbf{u}_k]_{i,j}} \quad \forall i,j \in  I \times J,
\end{equation}
where $\vect{b}(\vect{y}_k)$ is the likelihood vector, given $\vect{y}_k$, and calculated as
\begin{equation*}
\begin{aligned}
&\vect{b}(\vect{y}_k) \\
& = \text{vect}[\{\mathbb{P}[[\vect{Y}]_{ij}=[\vect{y}_k]_{ij}|[\vect{X}]_{ij}=m], \forall m,i,j \in  \mathcal{S} \times I \times J\}]   \\
&= \text{vect}[\{ [\mathbf{O}]_{m[\vect{y}_k]_{ij}}, \forall m,i,j \in  \mathcal{S} \times I \times J \}]  ,  
\end{aligned}
\end{equation*}
and $\mathbf{B}_k$ is a diagonal matrix with $\vect{b}_k$ as its diagonal.

(2) Update $\mathbf{u}_{k}$ using the state transition operator $\Phi(\cdot)$, which was defined in~\eqref{eq:state_transition_operator}, and the state estimate $\hat{\mathbf{p}}_k$ as 
\begin{equation}\label{eq:recursive_update}
    \mathbf{u}_{k+1} = \Phi \left(\hat{\mathbf{p}}_k\right).
\end{equation}

Using the full knowledge of the model, the Bayesian recursive state estimator can estimate the hidden state (green colored grids) as shown in Fig.~\ref{fig/state_estimation}.
 
\subsection{Online Parameter Estimation}\label{sec:SysID}
\begin{figure}[ht]
\vskip 0.0in
\begin{center}
\centerline{\includegraphics[width=\columnwidth]{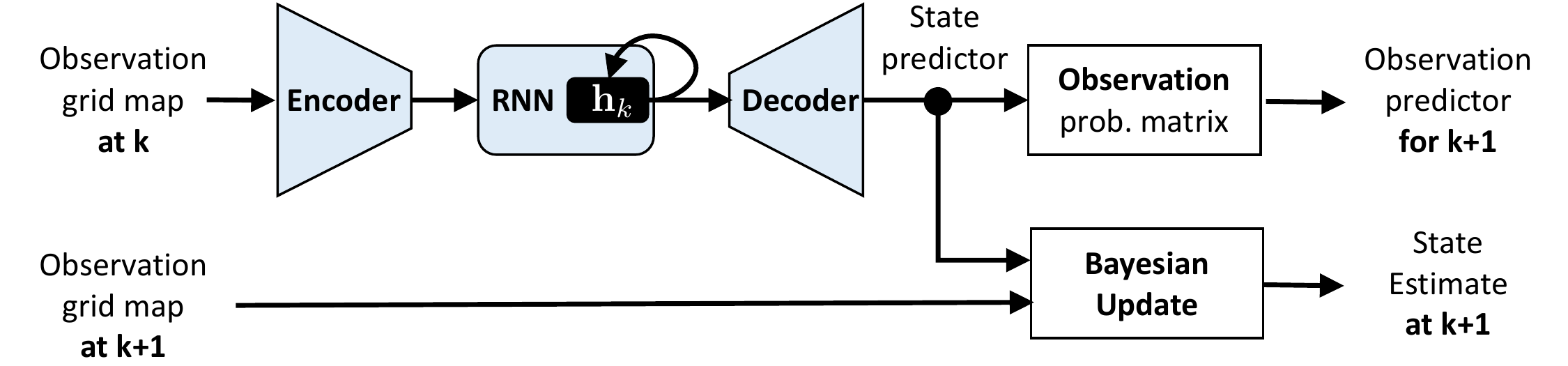}}
\caption{Spatiotemporal state estimation network. The dynamic autoencoder (colored light blue) replace the state predictor equation in~\eqref{eq:recursive_update} which is trained to predict the observation at $k+1$ using the observation at $k$ and the state of RNN denoted as $h_k$. The state estimate follows using the Bayesian update in~\eqref{eq:state_estimate_eqn}.}
\label{fig/diagram_dynamic_autoenc}. 
\end{center}
\vskip -0.2in
\end{figure}

The recursive HMM estimation~\cite{krishnamurthy2002recursive, yoon2019hidden} which uses a stochastic approximation~\cite{kushner2012stochastic} can be used to estimate parameters of the model. However, the estimation method~\cite{krishnamurthy2002recursive} becomes computationally intractable as we increase the size of the grid map. Hence, we employ the computational tools developed in deep learning that scales with a larger grid map when it uses GPUs. We approximate the recursive updates in~\eqref{eq:recursive_update} with the dynamic autoencoder network consisting of an encoder\footnote{The encoder compresses images into encodings~\cite{vincent2010stacked}.}, a recurrent neural network (RNN)\footnote{We use the gated recurrent unit (GRU) for the RNN component~\cite{cho2014learning}.}, and a decoder\footnote{The decoder generates a grid map, which is equivalently an $M$ channel image. The generator employs the deconvolutional layer structure in~\cite{radford2015unsupervised}.}, as illustrated in Fig.~\ref{fig/diagram_dynamic_autoenc}.

The deep neural networks in Fig.~\ref{fig/diagram_dynamic_autoenc} are trained to predict the future observation in the next time step given previous observations. In contrast to the HMM estimation methods~\cite{krishnamurthy2002recursive}, which only use the current observation, we use replay buffer~\cite{zhang2017deeper} to save recent trajectories to sample minibatch samples for training. The trajectory of the observation grid map with $|\mathcal{O}|$ possible observations can be seen as $|\mathcal{O}|$ channel image stream. The model in Fig.~\ref{fig/diagram_dynamic_autoenc} is trained to predict the observation image streams, i.e. observation grid map trajectories.

The system identification aims to obtain the parameter, which maximizes the likelihood of the observation given the state estimate calculated from the learned model. We maximize the likelihood by minimizing the cross-entropy loss between true image streams and the predicted image streams by a stochastic optimization, which samples trajectories saved in the replay buffer denoted $\mathcal{M}_\text{trajectory}$. The stochastic gradient for the optimization is calculated as follows:    
\begin{equation}\label{eq:sys_id}
    S^\text{sys}_n =  - \nabla_{\theta_\text{sys}} l^\text{sys}(\mathcal{D}_n^\text{trajectory};\theta_\text{sys}),
\end{equation}
where $\mathcal{D}_n^\text{trajectory}:=\{\{\mathbf{y}_k^1\}_{k=1}^K, \dots, \{\mathbf{y}_k^L\}_{k=1}^K\}$ denotes sampled $L$ trajectories from the rolling memory buffer $\mathcal{M}^\text{trajectory}$,  $\{\mathbf{y}_k^l\}_{k=1}^K$ denotes $l$\textsuperscript{th} sample image stream with length $K$, and $\theta_\text{sys}$ denotes the parameter of the model in Fig.~\ref{fig/diagram_dynamic_autoenc}. The loss function $l^\text{sys}(\cdot)$ is calculated as follows:
\begin{equation}\label{eq:loss_full_observation}
    l^\text{sys}(\mathcal{D}_n^\text{trajectory};\theta_\text{sys}) = \frac{1}{LK}\sum_{l=1}^L\sum_{k=1}^K H(\mathbf{y}_k^l, \hat{\mathbf{y}}_k^l),
\end{equation}
where
\begin{equation*}
    H(\mathbf{y}_k^l, \hat{\mathbf{y}}_k^l) = \sum_{m=1}^{|\mathcal{O}|}\sum_{i=1}^{W}\sum_{j=1}^{H} h([\mathbf{y}^l_k]_{m,i,j}, [\hat{\mathbf{y}}^l_k]_{m,i,j}),
\end{equation*}
and $\hat{\mathbf{y}}^l_k$ is the predicted observation grid map (or image frame) for the target $\mathbf{y}^l_k$, and $k$ denotes the time index, and the superscript $l$ denotes the sample index. Further, $i,j,m$ denote width, height, observation (or color) index for the image with width $W$, height $H$, number of possible observations (color channel), and $h(\cdot, \cdot)$ denotes binary cross entropy\footnote{For $y\in [0,1]$ and $\hat{y}\in (0,1)$, the binary cross entropy is calculated as $h(y, \hat{y})=y\log\hat{y} + (1-y)\log(1-\hat{y})$, and we follow the convention $0 = 0 \log 0$.} between observation probabilities (or pixel intensities) of $[\mathbf{y}^l_k]_{m,i,j} \in (0,1)$ and   $[\hat{\mathbf{y}}^l_k]_{m,i,j} \in [0,1]$. Each sample trajectory $\{\mathbf{y}_k\}$ is processed through the encoder, RNN, and the decoder with the following process. For the image stream sample $\{\mathbf{y}_1, \dots, \mathbf{y}_K\}$ and initial observation $\mathbf{y}_0$, we calculate
\begin{equation}\label{eq:dynautoenc}
    \begin{aligned}
    \vect{h}_{k+1} &= \text{RNN}(\vect{h}_k, \text{Encoder}(\vect{y}_k)), \quad \vect{h}_0 \sim \mathcal{N}(0,\mathbf{I})\\
    \hat{\vect{u}}_{k+1} & = \text{Decoder}(\vect{h}_{k+1})\\
    [\hat{\vect{y}}_{k+1}]_{i,j} & = \mathbf{O}^\top \, [\hat{\vect{u}}_{k+1}]_{i,j},
    \end{aligned}
\end{equation}
where $\hat{\vect{u}}_{k+1} \in \mathcal{P}^{|\mathcal{S}| \times W \times H}$ is the state predictor, which gets multiplied by the observation probability matrix $\mathbf{O}$, and we collect them into $\{\hat{\vect{y}}_1, \dots, \hat{\vect{y}}_K\}$.

The state predictor $\hat{\vect{u}}_{k}$ and current observation $\vect{y}_{k}$ are used to obtain the state estimate $\hat{\vect{p}}_{k}$ using the same Bayesian update in~\eqref{eq:state_estimate_eqn} as follows:
\begin{equation}\label{eq:state_estimate_eqn_dnn}
    [\hat{\mathbf{p}}_k]_{i,j} = \frac{\mathbf{B}([\mathbf{y}_k]_{i,j}) [\hat{\mathbf{u}}_k]_{i,j}}{\mathbf{b}([\mathbf{y}_k]_{i,j})^\top [\hat{\mathbf{u}}_k]_{i,j}} \quad \forall i,j \in  I \times J.
\end{equation}

\begin{figure}[ht]
\vskip 0.0in
\begin{center}
\centerline{\includegraphics[width=0.8\columnwidth]{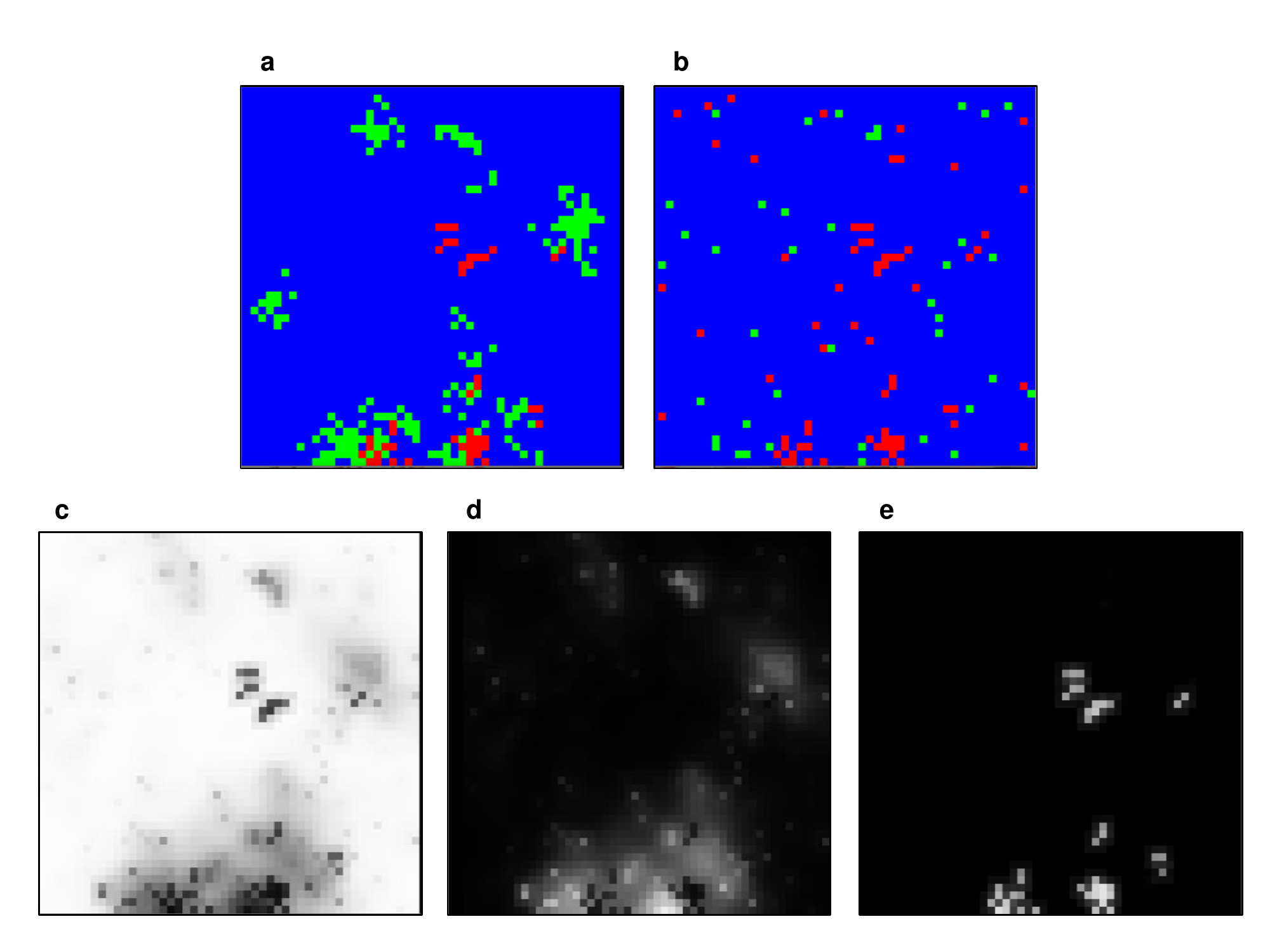}}
\caption{State estimation after online system identification. See the illustrative video
linked at \url{https://youtu.be/CZoAPa3Bya4}.}
\label{fig/param_state_estimation}
\end{center}
\vskip -0.2in
\end{figure}

After online adaptation (system identification) with thirty thousand time-steps (in time scale $k$), the state estimation network in~\eqref{fig/diagram_dynamic_autoenc} is learned to estimate the hidden state as shown in the observation grids of Fig.~\ref{fig/param_state_estimation}, and the 2\textsuperscript{nd} state grid map in Fig.~\ref{fig/param_state_estimation}d.

\subsection{Path Dependent Observation and Parameter Estimation}\label{sec:PathSysId}
For exploration using mobile robots (UAVs), the observations over the grid map depend on their trajectories. We can still use the same modeling framework in Fig.~\ref{fig/diagram_dynamic_autoenc}, which is  used by the dynamic autoencoder for predicting future observations, i.e. maximum likelihood estimation. The loss function in~\eqref{eq:loss_full_observation} is modified to consider the path-dependent observation as
\begin{equation}\label{eq:loss_mask_observation}
    l^\text{sys}(\mathcal{D}_n^\text{trajectory};\theta_\text{sys}) = \frac{1}{LK}\sum_{l=1}^L\sum_{k=1}^K H(\mathbf{y}_k^l \odot \mathbf{m}_k^l, \hat{\mathbf{y}}_k^l \odot \mathbf{m}_k^l),
\end{equation}
where $\mathbf{m}_k^l \in \mathcal{P}^{|\mathcal{S}| \times H \times W}$ denotes mask tensor that indicates grids visited by the robot at the slower time scale index $k$, i.e. when the robot visited the grid at $(i,j)$, then $[\mathbf{m}_k^l]_{i,j}=1$, and the other grids that the robot did not visit have zero values.

\begin{figure}[ht]
\vskip 0.0in
\begin{center}
\centerline{\includegraphics[width=0.8\columnwidth]{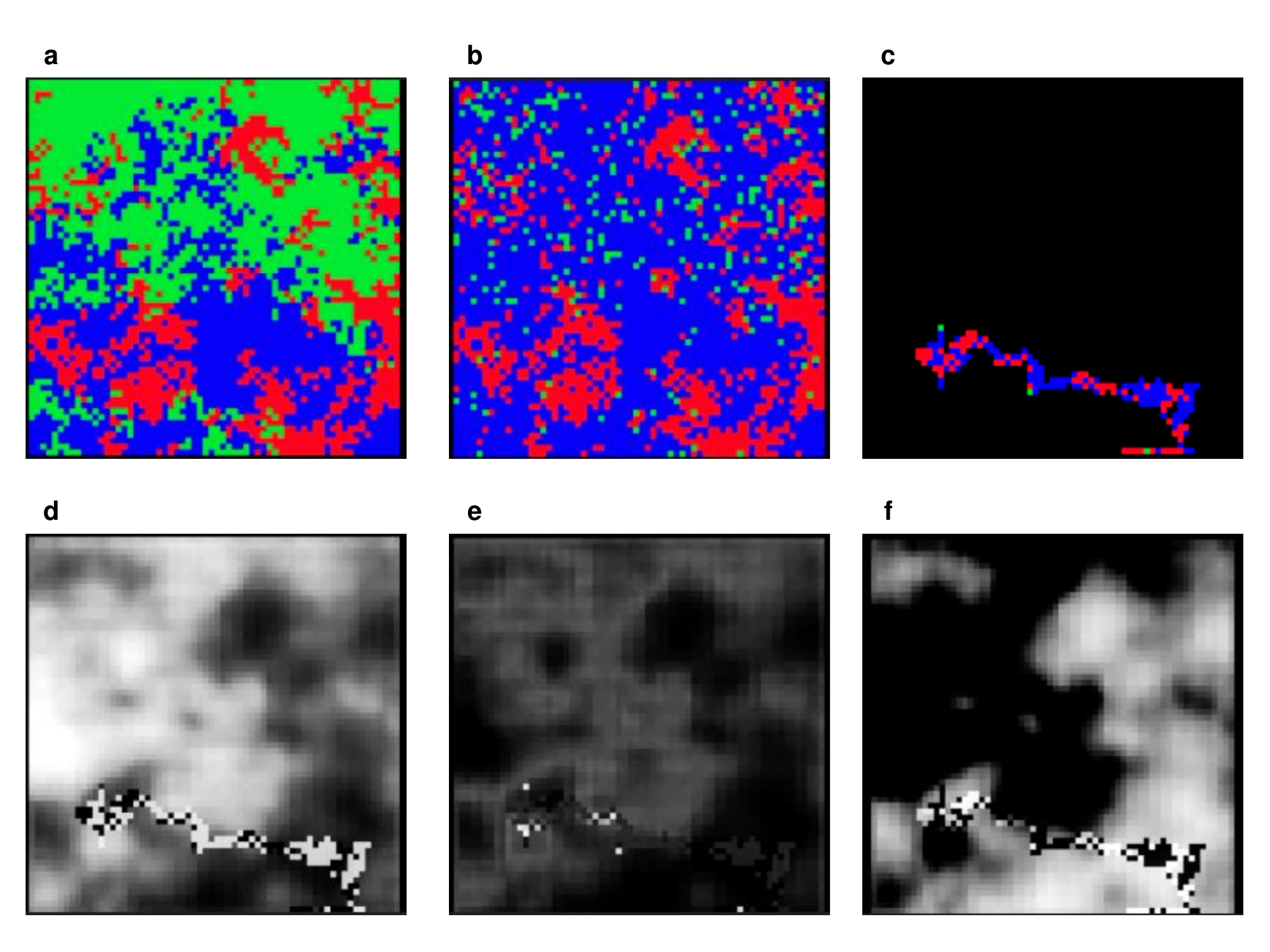}}
\caption{State estimation after online system identification: (a) True state map. (b) Observation map. (c) Observation map visited by the robot. Belief probability of (d) state 1 (blue); (e) state 2 (green); (f) state 3 (red). See the illustrative video linked at
\url{https://youtu.be/ej5Z90Pa2ys}.}
\label{fig/param_state_estimation_path_depend}
\end{center}
\vskip -0.2in
\end{figure}

As shown in Fig.~\ref{fig/param_state_estimation_path_depend}c, only grids of map visited by the mobile robot are observed and used for parameter and state estimation. Here, the robot follows a simple dynamic constraint, i.e., the vehicle can move only to its adjacent neighbor grids at a time. We assume that there are two different time scales: $k$ slower time scale (e.g., days) for estimation and $t$ faster time scale (e.g., minutes) for path planning.

\subsection{Short-Term Path Planning}\label{sec:planning}
The paths of the robot in Fig.~\ref{fig/param_state_estimation_path_depend} were not planned but randomly generated (random walks). The path can be planned to improve performance reward, given the state estimation grid map in Fig.~\ref{fig/param_state_estimation_path_depend}d, e, and f. For the path planning, we consider the following single integrator model for dynamic constraint as
\begin{equation}\label{eq:mpc_dynamic}
    \vect{z}_{t+1} = \vect{z}_{t} + \vect{v}_{t}, \quad \vect{z}_{0}\in I, \times J,
\end{equation}
where the position of the robot is denoted as $\vect{z}_t$, constrained within $I \times J$ and initialized with the ending coordinate of the previous path at $k-1$, and the velocity is denoted as $\vect{v}_t \in V$, and the set of velocity is  $V := \{(1,0), (-1,0), (0,1), (0,-1)\}$, i.e., the robot can move up, down, right, and left at a time.

We devise a cost function that depends on both the action $a_k$ chosen by the high-level decision-making agent (RL agent) and the uncertainties of the current state estimate as follows:
\begin{equation}\label{eq:mpc_running_cost}
    \tilde{c}(\vect{z}_{t}, \hat{\vect{p}}_k, a_t) = c_1(\vect{z}_{t}, \hat{\vect{p}}_k, a_t) +  \omega(a_t) \, c_2(\vect{z}_{t}, \hat{\vect{p}}_k),
\end{equation}
where $\hat{\vect{p}}_{k}$ is the state estimate grid map at time $k$ and $c_1(\cdot)$ denotes mission related costs, e.g., number of visits to grids estimated as green or red state. Here $c_2(\cdot)$ counts for the visit to grids with uncertain state estimate as
\begin{equation}\label{eq:uncertainty_cost}
    c_2(\vect{z}_{t}, \hat{\vect{p}}_k) = -\left(-\sum_{n=1}^N [\hat{\vect{p}}_k]_{n,i,j} \log [\hat{\vect{p}}_k]_{n,i,j}\right), \quad \vect{z}_{t} = (i,j),
\end{equation}
which calculates \emph{Shannon} entropy to promote the robot visiting the grids with greater uncertainties. The weight $\omega(a_t) \in [0, 1]$ in~\eqref{eq:mpc_running_cost} is the tuning parameter to consider the trade-off between exploitation and exploration.

The path planning aims to minimize the cumulative cost as 
\begin{equation}\label{eq:cost_cum_sum}
    \argmin_{\bar{\mathbf{v}}\in \bar{\mathbf{V}}} \frac{1}{T+1} \sum_{t=0}^{T}, \tilde{c}(\vect{z}_{t}, \hat{\vect{p}}_k, a_k),
\end{equation}
where $\bar{\mathbf{V}}$ denotes the possible set of velocity trajectories $\bar{\mathbf{v}}:=\{\mathbf{v}_0, \dots, \mathbf{v}_T\}$ with time-length of $T$.

We use sample based planning approaches by uniformly sampling from $V$ with the size $(N \times T)$, i.e. $N$ trajectories with length $T$. Then the $N$ trajectories of the velocities are integrated following the dynamic equation~\eqref{eq:mpc_dynamic} to generate position trajectories. The trajectories are then mapped to $N$ cost functions using~\eqref{eq:cost_cum_sum}. Finally, we choose the trajectory with the least cost as the outcome trajectory for the sample-based planning.


\subsection{Reinforcement Learning to Guide Short-Term Planning}\label{sec:RL}
\begin{figure}[ht]
\vskip -0.0in
\begin{center}
\centerline{\includegraphics[width=\columnwidth]{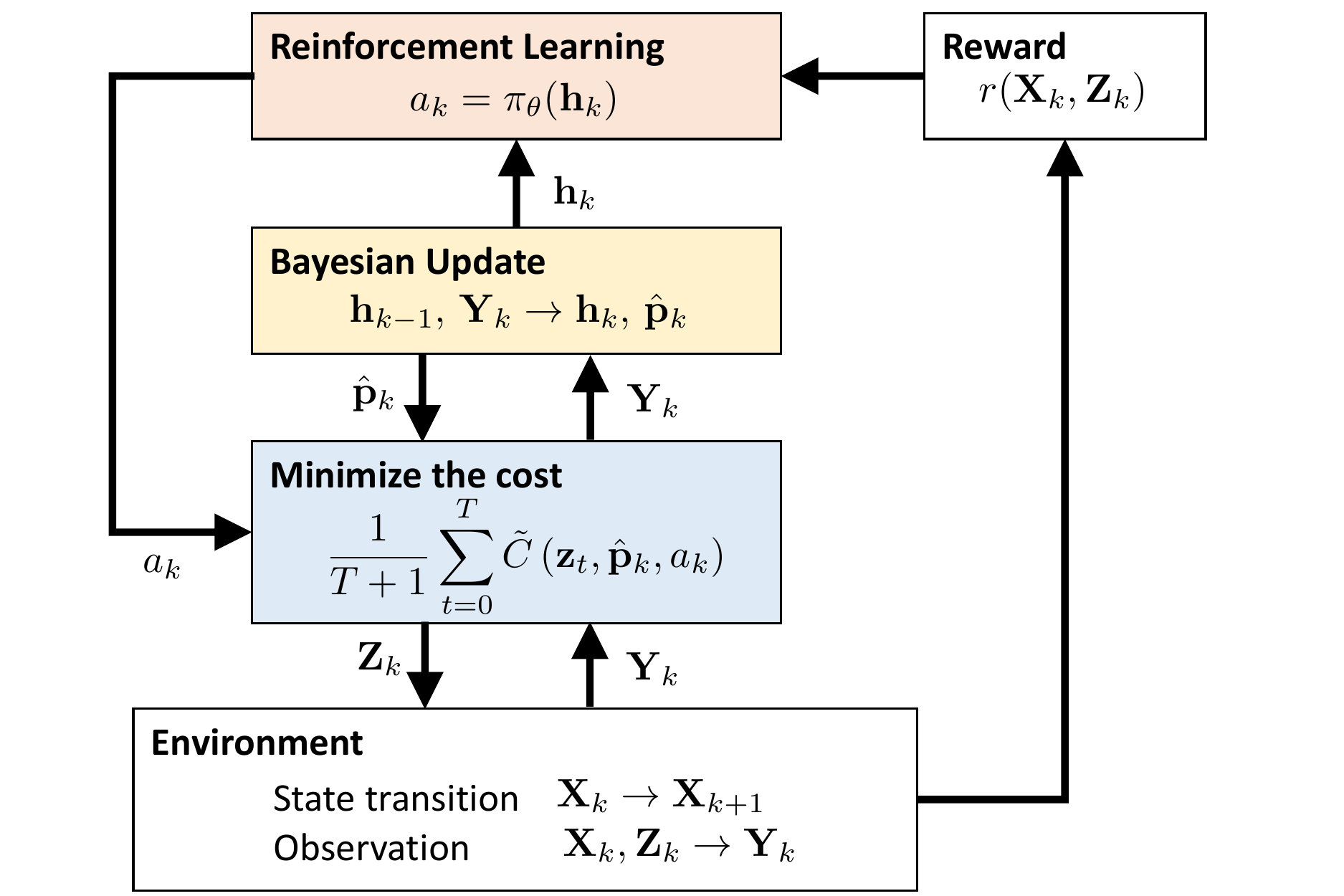}}
\caption{Adaptive path planning framework overview.}
\label{fig/diagram_rl}
\end{center}
\vskip -0.4in
\end{figure}
The high-level decision making $a_k$ at $k$ in the slow time scale is adaptive using a stochastic optimization (reinforcement learning). As in Fig.~\ref{fig/diagram_rl}, the policy $\pi_\theta(\cdot)$, given the encoding $\mathbf{h}_k$, chooses $a_k$, which is used as a cost parameter in~\eqref{eq:mpc_running_cost}. Here, $\mathbf{h}_k$ has information to calculate the state predictor $\hat{\vect{u}}_k$ and $\hat{\vect{p}}_k$ as in~\eqref{eq:dynautoenc} by saving all the previous observations, i.e., $\{\mathbf{y}_0, \mathbf{y}_1, \dots, \mathbf{y}_k\}$ through the recursive update of RNN in~\eqref{eq:dynautoenc}.

The high-level decision making deals with an uncertain environment modelled as a partially observable Markov decision process (POMDP). The POMDP  consists of the following state transition model $T(\cdot, \cdot)$ and observation model\footnote{Note that the observation $\mathbf{Y}$ depends on the trajectory $\mathbf{Z}$.}:
\begin{equation}\label{eq:pomdp}
\begin{aligned}
    T(\mathbf{X}, \mathbf{X}') &= \mathbb{P}[\mathbf{X}_{k+1}=\mathbf{X}' | \mathbf{X}_k=\mathbf{X} ] \\
    O(\mathbf{Y}, \mathbf{X}, \mathbf{Z}) &= \mathbb{P}[\mathbf{Y}_{k}=\mathbf{Y} | \mathbf{X}_k=\mathbf{X},  \mathbf{Z}_k=\mathbf{Z}],
\end{aligned}
\end{equation}
where $\mathbf{Z}_k = \{\mathbf{z}_0, \dots, \mathbf{z}_T\}$ denotes the trajectory of the robot determined by the short-term path planning.

We aim to improve the high-level decision making policy
\begin{equation}\label{eq:policy}
    a_k = \pi_{\theta}(\mathbf{h}_k)
\end{equation}
by searching for the optimal policy such that  
\begin{equation}\label{eq:optimal_policy}
     \theta_* = \argmax_{\theta\in \Theta}\mathbb{E}_{\pi_\theta}\left[ \sum_{k=0} ^ \infty \gamma^k r \left(\vect{X}_k,\vect{Z}_k(a_k, \hat{\mathbf{p}}_k)\right) \right],
\end{equation}
where the reward function $r(\cdot, \cdot)$ depends on both the state and the trajectory $\vect{Z}_k(a_k, \hat{\mathbf{p}}_k)$, and $\gamma \in (0, 1)$ denotes the discounting factor, and $\Theta$ is a set of feasible parameters for the policy, which generates the tuning parameter $a_k$ as actions of the reinforcement learning. We use $\vect{Z}_k(a_k, \hat{\mathbf{p}}_k)$ as the notation of the trajectory $\vect{Z}_k$ to explicitly show that the trajectory depends on both the state estimate $\hat{\vect{p}}_k$ and the high-level decision making $a_k$. The reward $r(\cdot, \cdot)$ in~\eqref{eq:optimal_policy} considers the main mission objective. For example, the number of visits to the grids which has the red colored state can be used as a reward. 
\begin{remark}
The reward function which needs hidden states is unrealistic. However, we can assume that there is a state which is observable with negligible uncertainties. For example, the red state in Fig.~\ref{fig/param_state_estimation_path_depend} is observed with low uncertainties.  
\end{remark}

We employ deep Q network (\emph{DQN})~\cite{mnih2015human} for learning the optimal policy. The \emph{DQN} and the online parameter estimation are integrated in forms of multi-time scale stochastic optimization~\cite{borkar1997stochastic}. Previously, the two-time scale algorithms were employed in reinforcement learning~\cite{konda2000actor} and generative adversarial networks~\cite{heusel2017gans}. The multi-time scale optimization has update iterations as follows:
\begin{equation}\label{eq:multi-time-scale}
\begin{aligned}
    \theta^\text{dqn}_{n+1} &= \theta^\text{dqn}_{n} + \epsilon_n^\text{dqn} S_n^\text{dqn}(\mathcal{D}_n^\text{transition})\\
    \theta^\text{sys}_{n+1} &= \theta^\text{sys}_{n} + \epsilon_n^\text{sys} S_n^\text{sys}(\mathcal{D}_n^\text{trajectory}),
\end{aligned}
\end{equation}
where the step sizes vanish  following the vanishing step size rules
\begin{equation}\label{eq:step-size-rule}
    \epsilon_n^\text{sys}/\epsilon_n^\text{dqn} \rightarrow 0 \quad
    \text{as} \quad n \rightarrow \infty,
\end{equation}
and $S_n^\text{sys}$ was defined in~\eqref{eq:sys_id}, and $S_n^\text{dqn}$ will be described in~\eqref{eq:dqn}. 

The system identification provides the state estimate information for the policy evaluation in \emph{DQN}. Hence, the update of the system identification is set to be slower than the \emph{DQN} update so that the policy evaluation can track the change of the state estimator of the system identification. In~\cite{heusel2017gans}, it was shown that the \emph{Adam}~\cite{kingma2014adam} step size rule can be set to implement the \textbf{two} \textbf{time} scale step size \textbf{update rule} (\emph{TTUR}) in~\eqref{eq:step-size-rule}. 

The system identification update $S_n^\text{sys}$ is defined in~\eqref{eq:sys_id} and we use the DQN update~\cite{mnih2015human} as
\begin{equation}\label{eq:dqn}
    S^\text{dqn}_n = - \nabla_{\theta_\text{dqn}} l^\text{dqn}(\mathcal{D}_n^\text{transition};\theta_\text{dqn})
\end{equation}
with the loss function as
\begin{equation*}
\begin{aligned}
    &l(\mathcal{D}_n^\text{transition};\theta_\text{dqn}) \\
    &=\frac{1}{M}\sum_{m=1}^M\left[ \left(  r_m + \gamma \max_{a'} Q(h_m',a'; \theta_n^-) - Q(h_m, a_m;\theta_n)  \right)^2 \right],
\end{aligned}
\end{equation*}
where the state transition samples $\mathcal{D}_n^\text{transition}=\{(h_1, a_1, r_1, h'_1), \dots, (h_M, a_M, r_M, h'_M)\}$ are randomly selected rows from the following replay buffer with size $\bar{L}$:
\begin{equation*}
\mathcal{M}_{\text{transition}} = 
\begin{bmatrix}
(h_{k-1-\bar{L}}, &a_{k-1-\bar{L}}, &r_{k-1-\bar{L}}, &h_{k-\bar{L}} )\\
 \vdots   &\vdots   &\vdots   &\vdots    \\
(h_{k-1}, &a_{k-1}, &r_{k-1}, &h_{k} )
\end{bmatrix}.
\end{equation*}

We summarise the entire procedure as the following multi-level stochastic optimization in Algorithm~\ref{alg:bi-level-optimization}.
\begin{algorithm}[ht]
   \caption{Multilevel Stochastic Optimization}
   \label{alg:bi-level-optimization}
\begin{algorithmic}
   \STATE {\bfseries Input:} State Estimation Network in Fig.~\ref{fig/diagram_dynamic_autoenc}
   \STATE {\bfseries \hspace{1cm}} with $\theta^\text{sys}_0$ of Encoder, RNN, and Decoder
   \STATE {\bfseries Input:} Deep Q network (DQN) with $\theta^\text{dqn}_0$.
   \STATE {\bfseries Input:} Spatiotemporal Model Environment.
   \STATE {\bfseries Input:} Replay buffers: $\mathcal{M}_\text{trajectory}$, $\mathcal{M}_\text{transition}$
   \STATE {\bfseries Output:} Fixed parameters: $\theta^\text{sys}_*$ and $\theta^\text{dqn}_*$
\FOR{$n=0$ {\bfseries to} $N_{\text{iterations}}$}
   \STATE Update RNN state $\mathbf{h}_k$ given $\mathbf{y}_k$ as in ~\eqref{eq:dynautoenc}
   \STATE $\quad \vect{h}_{k+1} \leftarrow \text{RNN}(\vect{h}_k, \text{Encoder}(\vect{y}_k))$
   \STATE $\quad \hat{\vect{u}}_{k} \leftarrow \text{Decoder}(\mathbf{h}_k)$
   \STATE Estimate the state using Bayes rule as in~\eqref{eq:state_estimate_eqn_dnn}
   \STATE $\quad \hat{\vect{p}}_{k} \leftarrow     \frac{\mathbf{B}(\mathbf{y}_k) \hat{\mathbf{u}}_k}{\mathbf{b}(\mathbf{y}_k)^\top \hat{\mathbf{u}}_k}$
   \STATE Use the DQN policy in~\eqref{eq:policy}
   \STATE $\quad a_k \leftarrow \pi_\theta(\mathbf{h}_k)$
   \STATE Generate a trajectory $\vect{Z}_k$ of the robot given $\hat{\vect{p}}_{k}$ and $a_k$ using the path planning with the cost in~\eqref{eq:cost_cum_sum}
   \begin{equation*}
    \argmin_{\bar{\mathbf{v}}\in \bar{\mathbf{V}}} \frac{1}{T+1} \sum_{t=0}^{T} \tilde{c}(\vect{z}_{t}, \hat{\vect{p}}_k, a_k)
   \end{equation*}
   \STATE Observe the environment given the trajectory and gets reward as in~\eqref{eq:pomdp} and~\eqref{eq:optimal_policy}
   \begin{equation*}
       \vect{X}_{k+1} \leftarrow  \vect{X}_k, \quad
       \vect{Y}_{k} \leftarrow \vect{X}_k, \, \vect{Z}_k,  \quad r_k \leftarrow r(\vect{X}_k, \, \vect{Z}_k)
   \end{equation*}
   \STATE Add new data to the replay buffers
   \STATE $\quad \mathcal{M}_\text{transition} \leftarrow (\vect{h}_{k-1}, a_{k-1}, r_{k-1}, \vect{k}_k)$
   \STATE $\quad \mathcal{M}_\text{trajectory} \leftarrow (\vect{y}_{k}, a_{k})$
   \STATE Update parameters with the gradients in~\eqref{eq:dqn} and~\eqref{eq:sys_id}
   \STATE $\quad \theta^\text{dqn}_{n+1} \leftarrow \theta^\text{dqn}_{n} + \epsilon_n^\text{dqn} S_n^\text{dqn}(\mathcal{D}^\text{transition}_n)$
   \STATE $\quad \theta^\text{sys}_{n+1} \leftarrow \theta^\text{sys}_{n} + \epsilon_n^\text{sys} S_n^\text{sys}(\mathcal{D}^\text{trajectory}_n)$
   \STATE Update the step sizes $\epsilon_n^\text{sys}$ and $\epsilon_n^\text{sys}$ using TTUR in~\cite{heusel2017gans}.
\ENDFOR
\STATE \textbf{Fix} the parameters with the current ones.
\end{algorithmic}
\end{algorithm}

\section{NUMERICAL EXAMPLE}\label{sec:numerical_examples}
\begin{figure}[h]
\vskip -0.0in
\begin{center}
\centerline{\includegraphics[width=0.8\columnwidth]{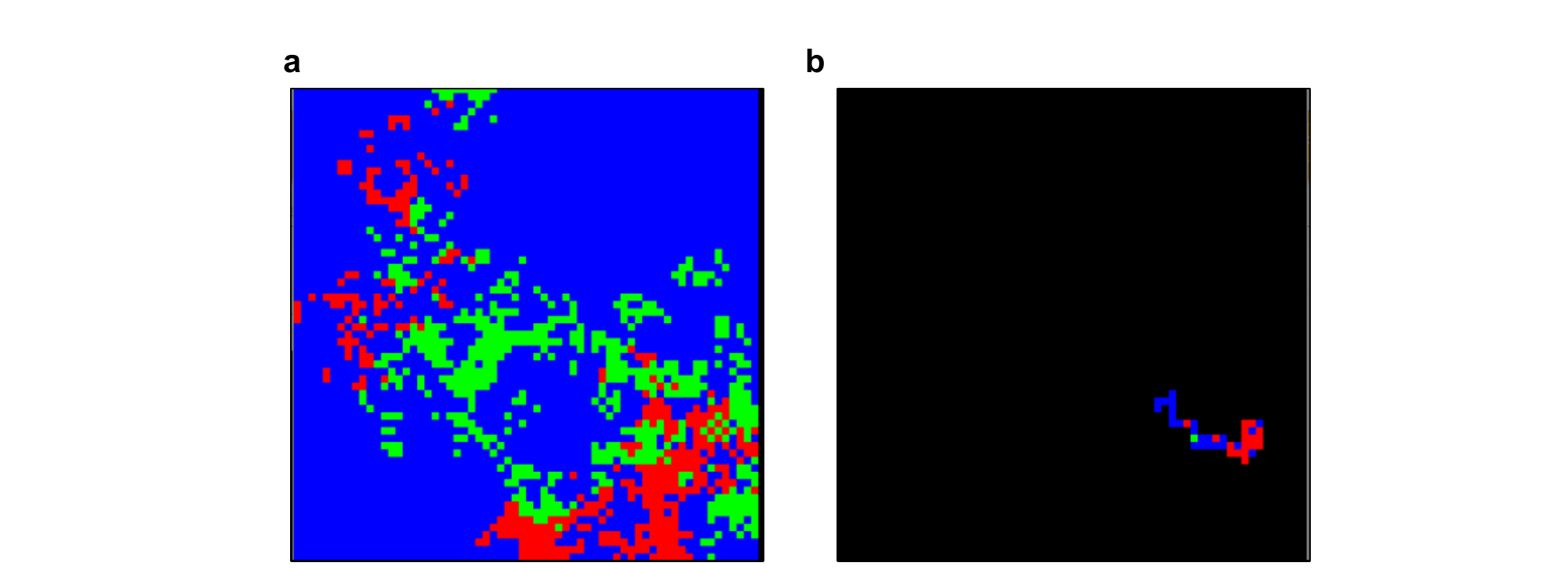}}
\caption{Incomplete observation by the robot on the mission to visit red grids: (a) Full state grid map. (b) Grids covered by the robot with a trajectory with time length $64$.}
\label{fig:mission}
\end{center}
\vskip -0.2in
\end{figure}

The simulation introduced in Fig.~\ref{fig/state_observation} was devised to model the spreading of wildfire. Each grid starts with a normal state (blue) and becomes a latent fire grid (green) and develops into fully grown fire (red), finally get extinguished to the normal state (blue). And the state transition dependence between the neighboring grids is biased along the vertical axis so that the fire spreading is directed downward as shown in the linked video within the caption of Fig.~\ref{fig/state_observation}. The parameter of the model and the implementation of Algorithm~\ref{alg:bi-level-optimization} can be found in the code repository linked at 
\url{https://github.com/stargaze221/WildFireModel}.

The goal of the path planning is to visit grids with fire (red) using the state estimate. During this mission, each trajectory takes $64$ time steps. Hence a trajectory can cover at most $1/64$ of the entire grid map ($64 \times 64$) as shown in Fig.~\ref{fig:mission}. Between trajectories along time $k$, the previous end position becomes the initial position in the next trajectory. The needs for state estimation in the simulation are due to path-dependent grid observation, latent fire, and dynamically spreading fire.

For the mission objective of visiting red grids, we specify the cost function in~\eqref{eq:mpc_running_cost} as
\vspace{-.1cm}
\begin{equation*}
    \tilde{c}(\vect{z}_{t}, \hat{\vect{p}}_k, a_k) = c_1(\vect{z}_{t}, \hat{\vect{p}}_k, a_k) +  \omega(a_k) \, c_2(\vect{z}_{t}, \hat{\vect{p}}_k),
\end{equation*}
where
$a_k \in \{0, 1, 2, 3\}$ and for $z_k = (i,j)$, we have
\begin{equation*}
    c_1(\vect{z}_{t}, \hat{\vect{p}}_k, a_k)= 
\begin{cases}
    - [\hat{\vect{p}}_k]_{0,i,j} & \text{if} \quad a_k = 0\\
    - [\hat{\vect{p}}_k]_{1,i,j} & \text{if} \quad a_k = 1\\
    - [\hat{\vect{p}}_k]_{2,i,j} & \text{if} \quad a_k = 2\\
       \quad  0                  & \text{otherwise}.
\end{cases}
\end{equation*}
The weight $\omega(a_k)$ to count for visiting uncertain grid is set as
\begin{equation*}
    \omega(a_k)= 
\begin{cases}
    1 & \text{if} \quad a_k = 3\\
    0 & \text{otherwise}.
\end{cases}
\end{equation*}
When we design the cost function above, we classify the actions into four classes: state 0 ($a_k=0$); state 1 ($a_k=1$); state 2 ($a_k=2$); uncertain grid ($a_k=3$) for the high-level decision making. The high-level decision-maker (RL agent) chooses either to  explore ($a_k=3$) or to exploit (visiting grids identified). In the cost design, we intended to promote disambiguation when the robot searches for fire in an uncertain environment. The reward for the mission objective is set to be the number of visits to fire grids on a path at $k$.

\begin{figure}[ht]
\vskip 0.0in
\begin{center}
\centerline{\includegraphics[width=\columnwidth]{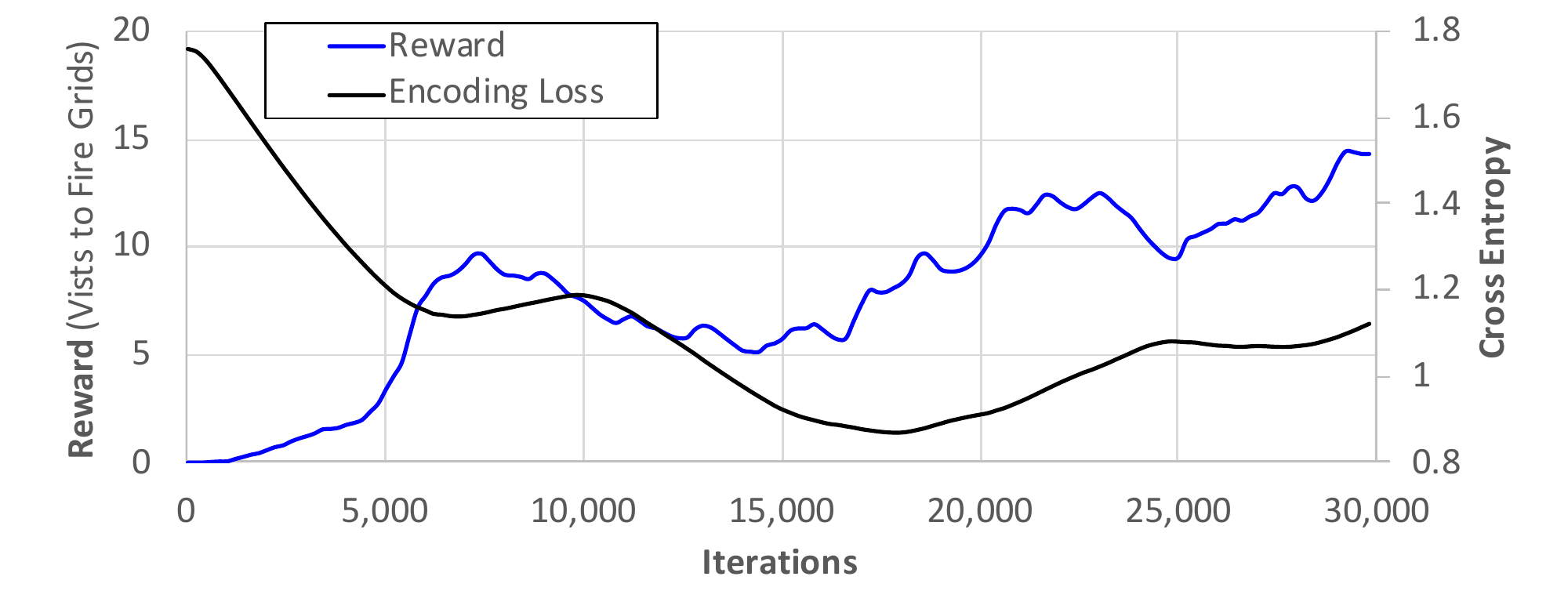}}
\caption{Performance improvement over iterations.}
\label{fig:sys_it_loss_iteraion}
\end{center}
\vskip -0.3in
\end{figure}

The multi-level stochastic optimization in Algorithm~\ref{alg:bi-level-optimization} improves the system identification loss in~\eqref{eq:sys_id} along the axis of the optimization iteration while improving the performance of finding fire as shown in Fig.~\ref{fig:sys_it_loss_iteraion}.

\begin{figure}[!ht]
\vskip 0.0in
\begin{center}
\centerline{\includegraphics[width=\columnwidth]{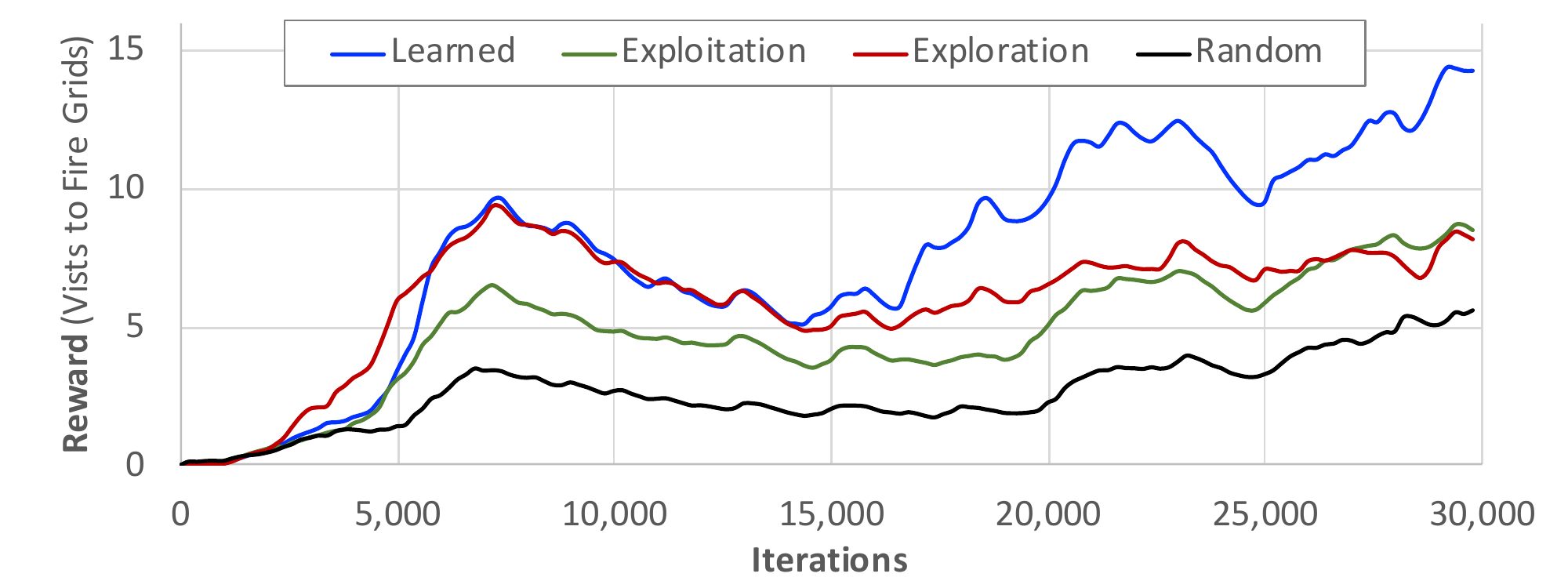}}
\caption{Performance comparison. See the illustrative video linked at \url{https://youtu.be/j5T2yRgK6bs}.}
\label{fig:policies_peformance_vs_iteraion}
\end{center}
\vskip -0.2in
\end{figure}

We compared the proposed method with random walks path and the \emph{greedy} policy (exploitation) which aims to visit the red state grids only by choosing the constant action matching to the red state. Compared to a random walk and the single-minded exploitation (or exploration), the path planning which uses the state estimation had better performance in visiting fire grids as in Table~\ref{tab:perform} and Fig.~\ref{fig:policies_peformance_vs_iteraion}, possibly due to the learned policy's ability to consider the trade-off between exploitation and exploration.
\begin{table}[ht]
\caption{Average rewards after 25,000 iterations.}
\label{tab:perform}
\vskip 0in
\begin{center}
\begin{small}
\begin{sc}
\begin{tabular}{lrc}
\toprule
Plan Method  &    Reward   &        \\
\midrule
Random Walk  &      4.3    & (baseline) \\
{\bf LEARNED POLICY}      &     12.6    &  293\% \\
Exploitation &      7.2    &  167\% \\
Exploratory  &      7.7    &  179\% \\
\bottomrule
\end{tabular}
\end{sc}
\end{small}
\end{center}
\vskip -0.1in
\end{table}

\section{CONCLUSION AND DISCUSSION}\label{sec:conclusion}
We presented an estimation and adaptive planning framework with a spatiotemporal model with incomplete state observations. We employed reinforcement learning (RL) to consider the trade-off between exploration and exploitation by integrating RL with the state estimator. Our framework is devised to address the two main challenges of unknown parameters and the partial state information due to noisy observation and path-dependent observations. As shown in the simulation, the proposed adaptive planner outperforms the greedy search that exploits the current state estimate. The adaptive planner has multiple components being updated at different time scales to ensure convergence by invoking the existing tools in stochastic approximations.

Future works include improving the short-term path planning to consider dynamic constraints specific to vehicles (wheeled vehicle or fixed-wing) and other mission costs (fuel efficiency). Also, designing the interface between the high-level decision-maker (RL agent) and low-level path planner (short-term path planner) needs to be theoretically analyzed, e.g., whether the designed action space spans the action space for the optimal policy of the associated POMDP.




\bibliographystyle{IEEEtran}
\bibliography{mybib}

\end{document}